\pdfoutput=1

\documentclass[11pt]{article}

\usepackage[preprint]{acl}

\usepackage{times}
\usepackage{latexsym}

\usepackage[T1]{fontenc}

\usepackage[utf8]{inputenc}

\usepackage{microtype}

\usepackage{inconsolata}

\usepackage{graphicx}

\usepackage{colorframed}  
\usepackage{xspace} 
\usepackage{colortbl} 
\usepackage{booktabs} 
\usepackage{amsmath} 
\usepackage{algorithm} 
\usepackage{algpseudocode} 

\newcommand{\struct}[1]{\texttt{\small #1}}
\newcommand{\utterance}[1]{\textit{#1}}
\newcommand{\phrase}[1]{\textit{``#1''}}

\newenvironment{Snugshade}[1][236,236,236]{
    \setlength{\itemsep}{0pt}
     \setlength{\parsep}{0pt}
     \setlength{\topsep}{0pt}
     \setlength{\partopsep}{0pt}
     \setlength{\leftmargin}{1.5em}
     \setlength{\labelwidth}{0em}
     \setlength{\labelsep}{0em} 
    \setlength{\parskip}{0pt}
    \definecolor{shadecolor}{RGB}{#1}%
    \begin{snugshade}
}{%
    \end{snugshade}%
}
\newcommand{\squishlist}{
    \begin{list}{$\bullet$}{ 
        \setlength{\itemsep}{0pt}
        \setlength{\parsep}{1pt}
        \setlength{\topsep}{1pt}
        \setlength{\partopsep}{0pt}
        \setlength{\leftmargin}{1.5em}
        \setlength{\labelwidth}{1em}
        \setlength{\labelsep}{0.5em} 
    } 
}
\newcommand{\squishend}{
  \end{list}  }
\newcommand{\myparagraph}[1]{\noindent \textbf{#1}.}
\newcommand{\myparagraphnospace}[1]{\noindent \textbf{#1}.}
\newcommand{\asqa}{\textsc{ASQA}\xspace}
\newcommand{\convmix}{\textsc{ConvMix}\xspace}

\newcommand{\dpr}{\textsc{DPR}\xspace}
\newcommand{\vicuna}{\textsc{Vicuna 13B}\xspace}

\newcommand{\silver}{\textsc{Silver}\xspace}

\title{Retrieving Contextual Information for Long-Form Question Answering using Weak Supervision}

\author{Philipp Christmann\thanks{Work was done during an internship at Amazon AGI.}
\\
  Max Planck Institute for Informatics \\
  \texttt{pchristm@mpi-inf.mpg.de} \\\And
  Svitlana Vakulenko \\
  Amazon AGI \\
  \texttt{svvakul@amazon.com} \\\AND
  Ionut Teodor Sorodoc \\
  Amazon AGI \\ 
  \texttt{csorionu@amazon.com} \\\And
  Bill Byrne \\
  Amazon AGI \\
  \texttt{willbyrn@amazon.co.uk} \\\And
  Adrià de Gispert \\
  Amazon AGI \\
  \texttt{agispert@amazon.com} \\}

\begin{document}

\maketitle
\begin{abstract}
Long-form question answering (LFQA) aims at generating in-depth answers to end-user questions, providing relevant information beyond the direct answer.
However, existing retrievers are typically optimized
towards information that directly targets the question,
missing out on such contextual information.
Furthermore, there is a lack of training data for relevant context.
To this end, we propose and compare different weak supervision techniques
to optimize retrieval for contextual information.
Experiments
demonstrate 
improvements on the end-to-end QA performance on ASQA,
a dataset for long-form question answering.
Importantly, as more contextual information is retrieved,
we improve the relevant page recall for LFQA by 14.7\%
and the groundedness of generated long-form answers
by 12.5\%.
Finally, we
show that long-form answers often anticipate likely
follow-up questions, via experiments on a conversational QA dataset.
\end{abstract}


\section{Introduction}

The goal of long-form question answering (LFQA) is to provide in-depth answers to end-user questions~\cite{stelmakh2022asqa, fan2019eli5}.
For example, for the user question 
\begin{Snugshade}
    \phrase{When did Lionel Messi start his career?}
\end{Snugshade}
a direct answer would be:
\begin{Snugshade}
    \struct{16 November 2003}
\end{Snugshade}
which is the date of his first team debut.

However, this answer naturally sparks a series of follow-up questions to obtain more contextual details~\cite{kumar2017incomplete}:
\begin{Snugshade}
    \phrase{For which club?}\\
    \indent \phrase{In which match?}\\
    \indent \phrase{What about his La Liga debut?}\\
    \indent \phrase{How did his career develop?}
\end{Snugshade}

A long-form answer aims to proactively supply a more complete and detailed response beyond the succinct direct answer, essentially anticipating such follow-up questions:
\begin{Snugshade}
\utterance{Lionel Messi started his career as a professional football player with FC Barcelona.
He made his first-team debut in a friendly against Porto on 16 November 2003, at the age of 16. 
\dots
}
\end{Snugshade}
We thus draw parallels between LFQA and conversational question answering (\mbox{ConvQA})~\cite{voskarides2020query, qu2020open, vakulenko2021question,christmann2023explainable,coman2023strong} and hypothesize that they can be treated as complementary tasks.
%

Existing approaches for LFQA based on retrieval-augmented generation~\cite{lewis2020retrieval, izacard2021leveraging, guu2020realm}
typically utilize retrievers that are optimized to obtain direct answers to questions.
Such retrieval systems thus often fail to retrieve relevant context, which is needed to generate faithful and comprehensive long-form answers.


\myparagraph{Approach}
We propose to train a specialised retriever for the task of LFQA that not only retrieves direct answers for a question, but also retrieves additional
context
required for grounding long-form answers.
Our goal is to retrieve both direct answers and contextual information in one shot.

The major bottleneck for training a retriever for LFQA is the absence of training data.
LFQA datasets~\cite{fan2019eli5, stelmakh2022asqa} contain questions and long-form answers but not the ground-truth passages required to produce those long-form answers.
In this work, we propose
a mechanism to automatically infer
\textit{silver passages},
designed to ground both~(i) the direct answers, \textbf{and}
(ii) the contextual information.
These passages should provide sufficient evidence to support information in the long-form answer.
This is different from previous work
on factoid short-form QA,
which identified such passages by matching only against the
direct answers~\cite{shen2023neural}.

Based on these silver passages, we train BERT-based re-ranking models~\cite{nogueira2019passage}.
These re-rankers are applied to the
initial retrieval results,
to enhance recall of contextual information in the top-ranked passages.
These top-ranked passages are
then provided as input
to an LLM
to generate the long-form answer.

We conduct experiments on \asqa~\cite{stelmakh2022asqa}, a dataset for LFQA,
and show that we substantially improve end-to-end QA performance, while also increasing the
groundedness of the long-form answer w.r.t. the retrieved passages.



Experiments on the ConvQA dataset \convmix~\cite{christmann2022conversational}
demonstrate that
our method generates long-form answers that often also contain answers to the follow-up questions,
when provided only with the first question of the conversation as input.
This indicates that LFQA can indeed anticipate
likely follow-up questions.


\vspace*{0.2cm}
\myparagraph{Contributions}
\squishlist
    \item A novel mechanism using the target long-form answers to identify silver passages expressing relevant contextual information.

    
    \item Improving end-to-end QA performance, achieving state-of-the-art performance on the competitive \asqa benchmark.
    
    \item An investigation into the relationship between the LFQA and ConvQA tasks as alternatives for satisfying the same information needs.
\squishend


\section{Identifying silver passages}
\label{sec:silver}


The key idea for obtaining such silver passages is to utilize both the long-form answers (LFAs) \textbf{and} direct answers (DAs), as annotated in the \asqa dataset, jointly as a weak supervision signal for passage relevance.
We retrieve a large set of passages first (say $100$), using
first-stage retrieval
~\cite{karpukhin2020dense},
and then choose
up to $k$
silver passages from this candidate pool.

We considered three techniques
for matching candidate passages against an LFA:
(i)~lexical matching, (ii)~semantic similarity, and (iii)~LLM perplexity.
Our proposed approach matches against a combination of both LFAs and DAs, and we also compare these against matching only with DAs.

\myparagraph{Lexical matching with~LFA}
We initially evaluated (i) token recall, (ii) Jaccard similarity between token sets, and (iii) ROUGE-L~\cite{lin2004rouge},
and found that plain token recall works best.
We thus compute the matching score for a pair of a candidate passage $p$ and the $LFA$ as:
\begin{equation}
    match(p, LFA) = \frac{|tokens(p) \cap tokens(LFA)|}{|tokens(LFA)|}
\end{equation}
where $tokens$ are sets of words produced by a tokenizer with stopword removal.

\myparagraph{Semantic similarity with~LFA}
We use a pre-trained Sentence-Transformer model\footnote{\url{https://huggingface.co/sentence-transformers/nli-roberta-base-v2}}~\cite{reimers2019sentence, vaswani2017attention},
to compute the semantic similarity
between a candidate $p$ and the $LFA$ as follows
(where $Enc$ is the text encoder, and $\cdot$ is the dot product):
\begin{multline}
    match(p, LFA) = Enc(p) \cdot Enc(LFA)
\end{multline}

\myparagraph{LLM perplexity of~LFA}
Inspired by the approach used in Toolformer~\cite{schick2023toolformer},
we compute the LLM perplexity of the target $LFA$ (of length $n$), given a candidate passage $p$, as follows:
\begin{multline}
    match(p, LFA) \\ = 
     - \sum\nolimits^{n}_{j=1} \log P(t_j | C, p, t_1...t_{j-1})
\end{multline}
where $P$ is the probability function of the LLM, $t_j$ is the $j$-th token of the LFA,
and $C$ is the same prompt as the one applied during LLM training and inference.
$C$ includes a random sample of $k$-1 candidate passages.
We observed that adding this random sample substantially improves the performance as it makes the samples closer to the input seen during LLM inference/training, i.e. a context with $k$ passages.

\myparagraph{Matching with DA}
All candidate passages that contain one of the DAs are considered as relevant.
Here we consider exact lexical matches since the DAs are relatively short.
This is the typical approach when the goal is to provide crisp answers~\cite{shen2023neural}.
Since there might be multiple passages matching the answer, and answers may also be matched out-of-context,
we sort all answer-matching passages by their token-recall with the question.

\myparagraph{\silver: Matching with LFA \& DA}
Finally, we consider the combination of matching both the LFA and the set of DAs, for selecting $k$ silver passages.
First, we ensure that each DA is matched by at least one of the candidate passages.
In case there are multiple candidate passages matching the same DA, the one with the highest matching score with the LFA is
chosen as relevant.
The remaining passages are chosen based on their LFA matching score,
to obtain a total of $k$ silver passages.
We utilize lexical matching with the LFA, which showed strong results (see Sec.~\ref{sec:experiments}) and is computationally inexpensive.
This combined variant is the \silver approach proposed in this work, as it jointly optimizes towards information for directly answering the question and relevant contextual information.

\vspace*{0.2cm}
\myparagraph{Pseudo-code for \silver approach}
Algorithm~\ref{alg:silver} in the Appendix illustrates the end-to-end workflow of deriving silver passages, fine-tuning the re-ranker, and fine-tuning the LLM on LFQA data.

\begin{table*}
    \newcolumntype{G}{>{\columncolor [gray] {0.90}}c}  
    \resizebox*{\textwidth}{!}{
        \begin{tabular}{l G G | G G G G c c | c c}
            \toprule
                \textbf{}      & \multicolumn{6}{G}{\textbf{\asqa~\cite{stelmakh2022asqa}}}  & \multicolumn{4}{c}{\textbf{\convmix~\cite{christmann2022conversational}}} \\
            \midrule
                \textbf{Metric} $\rightarrow$
                & \textbf{Recall:} & \textbf{Recall:} &  \textbf{} & \textbf{}   &  \textbf{} & \textbf{}
                & \textbf{Recall:} & \textbf{Recall:}  & \textbf{}  & \textbf{} \\
                
                \textbf{Retrieval Method} $\downarrow$
                & \textbf{Direct Ans.} & \textbf{Wikipage} & \textbf{Ground.} & \textbf{ROUGE-L} & \textbf{D-F1} & \textbf{DR}
                & \textbf{Direct Ans.} & \textbf{Follow-up Ans.}  & \textbf{Ground.}  & \textbf{C-F1} \\
            \midrule
                \textbf{Baseline retriever}
                &	$0.489$	 &	$0.450$	 &	$0.763$   &	$42.5$ 	&	$27.7$  &	$34.3$
                &	$0.558$  &	$0.260$ &	$0.726$   &	$20.2$  \\

                \textbf{+ pre-trained re-ranker}
                &	$0.336$	 &	$0.345$	 &	$0.664$   &	$40.1$ 	&	$23.7$  &	$30.8$
                & $0.514$ & $0.235$
                & $0.683$ & $19.9$ \\

            \midrule

                \textbf{Matching with~DA}
                &	$\mathbf{0.498}$	 &	$0.483$	 &	$0.810$   &	$42.9$ 	&	$28.0$  &	$34.7$
                &	$0.622$  &	$0.292$ &	$0.803$   &	$\mathbf{20.8}$  \\

            \midrule
                \textbf{Lexical matching with~LFA} 
                &	$0.489$	 &	$0.501$	 &	$0.851$   &	$43.3$ 	&	$28.3$  &	$35.0$
                &	$\mathbf{0.635}$  &	$\mathbf{0.306}$  &	$0.828$  &	$\mathbf{20.8}$  \\

                \textbf{Semantic similarity with~LFA} 
                &	$0.482$	 &	$0.500$	 &	$0.829$   &	$42.8$ 	&	$27.2$  &	$34.1$
                &	$0.619$  &	$0.303$ &	$0.817$   &	$20.4$  \\

                \textbf{LLM perplexity of LFA} 
                &	$0.483$	 &	$0.495$	 &	$0.845$   &	$43.3$ 	&	$26.7$  &	$34.0$
                &	$0.625$  &	$0.305$  &	$0.838$  &	$20.7$  \\
            
            \midrule
                \textbf{\silver: Matching with~LFA \& DA} 
                &	$0.491$	 &	$\mathbf{0.516}$	 &	$\mathbf{0.858}$   &	$\mathbf{43.4}$ 	&	$\mathbf{29.0}$  &	$\mathbf{35.5}$
                &	$\mathbf{0.635}$  &	$\mathbf{0.306}$ &	$\mathbf{0.839}$   &	$\mathbf{20.8}$  \\            
                
            \bottomrule
        \end{tabular} 
    }
    \vspace*{-0.2cm}
    \caption{Main results comparing retrieval and end-to-end QA performance on \asqa and \convmix. We use \dpr~\cite{karpukhin2020dense} as our retrieval baseline, which has been commonly used on the \asqa dataset.}
    \label{tab:main-res}
    \vspace*{-0.1cm}
\end{table*}

\begin{table}
    \newcolumntype{G}{>{\columncolor [gray] {0.90}}c}  
    \resizebox*{\columnwidth}{!}{
        \begin{tabular}{l G G G c}
            \toprule
                \textbf{Method}
                & \textbf{Length} & \textbf{ROUGE-L} & \textbf{D-F1} & \textbf{DR} \\
            \midrule
                \textbf{FLARE~\cite{jiang2023active}}
                & $-$ & $34.3$ &	$28.2$   &  $31.1$  \\
                \textbf{PaLM 540B~\cite{amplayo2022query}}
                & $64.1$ & $40.7$  &	$27.8$  &	$33.5$  \\
                \textbf{JPR + T5-large~\cite{stelmakh2022asqa}}
                & $71.6$ & $43.0$  &	$26.4$  &	$33.7$  \\
                \textbf{SIXPAQ~\cite{sun2023answering}}
                & $63.5$ &	$43.8$ &	$28.9$   &  $35.6$  \\
            \midrule
                \textbf{\dpr + \vicuna}
                & $70.5$ & $43.0$  &	$28.3$  &	$34.9$  \\
                \textbf{\silver + \vicuna (proposed)} 
                & $70.0$ &	$\mathbf{44.1}$ &	$\mathbf{30.8}$   &  $\mathbf{36.9}$  \\
            \bottomrule
        \end{tabular}
    }
    \vspace*{-0.2cm}
    \caption{Comparison with state-of-the-art on \asqa.}
    \label{tab:asqa-sota}
    \vspace*{-0.3cm}
\end{table}


\begin{table*} [t]
\centering
    \resizebox*{\textwidth}{!}{
        \begin{tabular}{p{4.2cm} | p{18cm}}
            \toprule
                \textbf{Question} 
                    & \phrase{Who played grandpa joe in charlie and the chocolate factory?} (from \asqa) \\
            \midrule
                \textbf{{\dpr + \vicuna}}
                    & \utterance{Willy Wonka and the Chocolate Factory is a {\color{orange}{{\color{orange}{1970}}}} children's book by {\color{orange}{Roald}} {\color{orange}{Dahl}}. It has been {\color{orange}{adapted}} into {\color{orange}{various}} {\color{orange}{media}}, {\color{orange}{including}} a 1971 film and a 2005 film. In the 1971 film, Grandpa Joe is {\color{orange}{played}} by Wilbur {\color{orange}{Robinson}}. In the 2005 film, Grandpa Joe is {\color{orange}{played}} by David Morris.} -- \textbf{Groundedness: $\mathbf{0.61}$} \\
            \midrule
                \textbf{{\silver + \vicuna} \newline(proposed)}
                    & \utterance{Charlie and the Chocolate Factory is a 2005 musical fantasy comedy film directed by Tim Burton and written by John August, based on the 1964 British novel of the same name by Roald Dahl. The film stars Johnny Depp as Willy Wonka and Freddie Highmore as Charlie Bucket. The storyline follows Charlie, who wins a contest and, along with four other contest winners, is led by Wonka on a tour of his chocolate factory, the most magnificent in the world. In the film, Grandpa Joe is played by David Kelly. In the {\color{orange}{{\color{orange}{2010}}}} musical adaptation of Charlie and the Chocolate Factory, Grandpa Joe is played by {\color{orange}{Richard}} {\color{orange}{Dreyfuss}}.} -- \textbf{Groundedness: $\mathbf{0.94}$}\\
            \toprule
                \textbf{Question} 
                    & \phrase{Number of episodes in The Sopranos?} (from \convmix) \\
            \midrule
                \textbf{{\dpr + \vicuna}}
                    & \utterance{The Sopranos is an {\color{orange}{American}} {\color{orange}{crime}} {\color{orange}{drama}} television series {\color{orange}{created}} by David Chase. The series {\color{orange}{ran}} for {\color{orange}{73}} episodes over six seasons, from {\color{orange}{January}} {\color{orange}{15}}, {\color{orange}{1999}}, to {\color{orange}{June}} {\color{orange}{11}}, {\color{orange}{2007}}}. -- \textbf{Groundedness: $\mathbf{0.23}$} \\
            \midrule
                \textbf{{\silver + \vicuna} \newline(proposed)}
                    & \utterance{The Sopranos is an American crime drama television series created by David Chase. The series portrays the difficulties that fictional New Jersey-based, Italian American mobster Tony Soprano (James Gandolfini) faces as he tries to balance his home life and his criminal organization. The series features Tony's family members, mafia colleagues, and rivals in prominent roles and story arcs, most notably his wife Carmela (Edie Falco) and protégé Christopher Moltisanti (Michael Imperioli). The series ran for 86 episodes over six seasons.} -- \textbf{Groundedness: $\mathbf{1.00}$} \\
            \bottomrule
        \end{tabular}
    }
    \vspace*{-0.1cm}
    \caption{Anecdotal examples from both datasets (randomly sampled) illustrating the benefits of explicitly retrieving contextual information. Un-grounded tokens (i.e., the ones not appearing in the retrieved passages) are highlighted in {orange}. The fraction of grounded tokens is greatly improved with our proposed approach in both cases.}
    \label{tab:anecdotes}
\end{table*}

\section{Experiments}
\label{sec:experiments}


\myparagraphnospace{Benchmarks}
We conduct experiments on two datasets:
(i) {\asqa}~\cite{stelmakh2022asqa}, a dataset for LFQA,
and (ii) {\convmix}~\cite{christmann2022conversational}, a dataset for ConvQA.
We use \asqa for fine-tuning
the re-rankers and LLMs.
Experiments on \convmix use the same models,
thus also test the generalizability of the approach.

\myparagraph{Retrieval metrics}
For evaluating retrieval, we measure recall in the top-$5$ retrieved passages.
On \asqa, we compute \textit{Direct Answer Recall} as the fraction of DAs 
appearing in the retrieved passages.
As a proxy for recall of contextual information, 
we measure \textit{Wikipage Recall} as the fraction of the ground-truth relevant Wikipages matched with our retrieval results.
A Wikipage is matched if we retrieve a passage from the respective page.
On \convmix, we measure \textit{Direct Answer Recall} / \textit{Follow-up Answer Recall} as the fraction of DAs for the
first question / follow-up questions appearing in the retrieved passages.

\myparagraph{Metrics}
On \asqa, we keep the metrics used in the original work,
and use their evaluation code\footnote{\url{https://github.com/google-research/language/tree/master/language/asqa}}.
This includes \textit{ROUGE-L}~\cite{lin2004rouge} for measuring the overlap of the generated text with one of the two reference LFAs.
For answer correctness, the Disambig-F1 (\textit{D-F1}) metric is used, measuring the fraction of question interpretations answerable from the generated LFA.
A pre-trained machine reading comprehension
(MRC)
~\cite{hermann2015teaching} model is used to this end.
The \textit{DR} metric combines the two metrics via the geometric mean.

For \convmix, inspired by the D-F1 metric,
we compute
\textit{C-F1} as the fraction of conversational questions
answerable from the generated LFA.
We use the same MRC model as for the D-F1 metric.

In addition, we measure \textit{Groundedness} as the fraction of tokens
in the answer that is also present in the retrieved passages.
Similar to lexical matching in Sec.~\ref{sec:silver}, stopwords are not considered.

\myparagraph{Configuration}
 We use \dpr~\cite{karpukhin2020dense}\footnote{DPR is trained on NQ~\cite{kwiatkowski2019natural} and used along the corresponding Wikipedia dump.} for first-stage retrieval, which was shown to outperform \textsc{Bm25}~\cite{robertson2009probabilistic}
on \asqa in previous work~\cite{sun2023answering},
and is still considered state-of-the-art on NaturalQuestions~\cite{kwiatkowski2019natural}, which is a superset of ASQA.

We use Vicuna 13B 1.5~\cite{zheng2023judging} as LLM for generating LFQAs and fine-tune it for $1$ epoch using the long-form answers in the \asqa dataset as the target output.
The input to the model are the question and the top-$5$ passages retrieved either by \dpr or by our \silver re-rankers.
We use the same
LLM
prompt as in the original \asqa paper to combine the question and the retrieved passages
for consistency.

Further details
on the
setup
in Appendix~\ref{app:setup}.

\subsection{Results}

\myparagraphnospace{Recall of contextual information is enhanced}
Table~\ref{tab:main-res} shows our main results.
The foremost take-away is that recall of contextual information is greatly improved
when incorporating our \silver re-rankers, compared to \dpr.
On \asqa, Wikipage recall increases from $0.450$ to $0.516$.
On \convmix, follow-up answer recall is improved from $0.260$ to $0.306$,
indicating that our \silver re-rankers aid the LLM to anticipate and successfully answer follow-up questions.

\myparagraph{Lexical matching shows strong performance}
An interesting finding is that simple lexical matching achieves better performance (DR of $35.0$) compared to the more
complex and computationally expensive variants based on semantic similarity (DR of $34.1$) or LLM perplexity (DR of $34.0$).
This result demonstrates that lexical matching is sufficiently robust
for the relatively long
answers.

\myparagraph{Improving end-to-end QA performance}
Combining LFA and DA to provide supervision signal for training the re-ranker leads to the best performance (DR of $35.5$),
substantially improving over the baseline results with \dpr (DR of $34.3$).
Both improvements in answer formulation (ROUGE-L of $43.4$ vs. $42.5$)
and provision of the right answer in an appropriate context (D-F1 of $29.0$ vs. $27.7$)
contribute to this overall increase in performance.

\myparagraph{Groundedness is substantially enhanced}
Another key take-away is the effect of our \silver re-rankers on the groundedness of the generated answers.
The groundedness is
dramatically
improved on both datasets compared to \dpr retrieval
($0.763$ to $0.858$ on \asqa and $0.726$ to $0.839$ on \convmix).
This result indicates that our generated LFAs are more likely to be based on the retrieved passages rather than hallucinated by the LLM.

\myparagraph{Comparison against a pre-trained re-ranker}
We also conducted an experiment with a pre-trained re-ranker replacing our proposed \silver re-rankers (results are shown in Table~\ref{tab:main-res}).
We used the same Sentence-Transformer as for our semantic similarity variant\footnote{\url{https://huggingface.co/sentence-transformers/nli-roberta-base-v2}}.
As can be expected, recall drops substantially, which leads to a much lower DR score.
Interestingly, since relevant information is missing from the retrieval results,
the groundedness of answers is greatly reduced in comparison to our proposed approach.
Note that our re-rankers, once trained on the \asqa dataset, can be successfully applied to a different dataset (\convmix in our experiments).

\subsection{Analysis}
\myparagraphnospace{Anecdotal examples}
Table~\ref{tab:anecdotes} 
demonstrates how our approach can improve the groundedness of generated answers:
the fraction of un-grounded tokens (i.e. the ones not present in the retrieved passages), as depicted in orange,
is much higher when conditioned on \dpr retrieval.
When conditioned on our \silver retrieval,
the LLM can mostly rely on the information in the provided passages for generating LFAs.
This can reduce factual hallucinations compared to the DPR-based variant, as illustrated in the second example:
as DPR retrieval is insufficient, the LLM hallucinates
incorrect contextual information (e.g., that Sopranos started on \textit{\color{orange}{January 15, 1999}}).

\myparagraph{Comparison with state-of-the-art}
Our approach achieves state-of-the-art performance on the \asqa dataset, as shown in Table~\ref{tab:asqa-sota}.

\begin{figure} [t]
    \centering
     \includegraphics[width=\columnwidth]{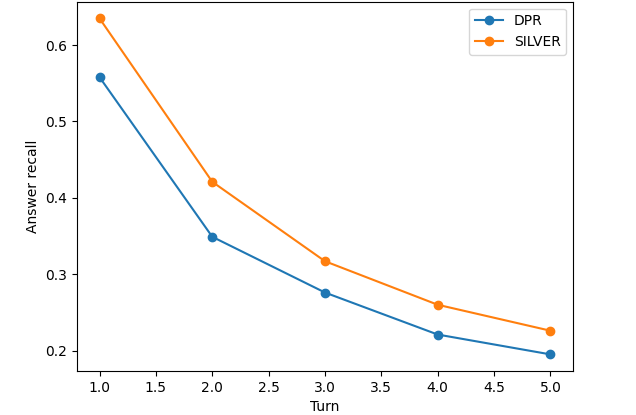}
     \vspace*{-0.7cm}
     \caption{Answer recall per turn on \convmix.}
     \label{fig:recall-per-turn}
\end{figure}

\myparagraph{Answer recall per turn}
We plot the average answer recall per turn on \convmix in Fig.~\ref{fig:recall-per-turn}.
As expected,
the answer recall drops
as the conversation drifts away from the initial topic.
Recall with our \silver re-ranker
remains consistently higher than for \dpr.


\begin{table} \small
    \newcolumntype{G}{>{\columncolor [gray] {0.90}}c}  
    \resizebox*{\columnwidth}{!}{
        \begin{tabular}{l G G G c}
            \toprule
                \textbf{Method}  & \textbf{ROUGE-L} & \textbf{D-F1} & \textbf{DR} \\
            \midrule
                \textbf{\dpr + \textsc{Vicuna 7B}}  & $40.6$ &	$26.0$   &  $32.5$  \\
                \textbf{\dpr + \vicuna}             & $42.5$ &	$27.7$   &  $34.3$  \\
                \textbf{\dpr + \textsc{Vicuna 33B}} & $42.6$ &	$27.8$   &  $34.3$  \\
            \midrule
                \textbf{\silver + \textsc{Vicuna 7B}}   & $42.3$ &	$26.3$   &  $33.3$  \\
                \textbf{\silver + \vicuna}              & $43.4$ &	$29.0$   &  $35.5$  \\
                \textbf{\silver + \textsc{Vicuna 33B}}  & $43.6$ &	$29.0$   &  $35.6$  \\
            \bottomrule
        \end{tabular}
    }
    \vspace*{-0.2cm}
    \caption{Comparison of end-to-end QA performance with different model sizes of the \textsc{Vicuna} model family.}
    \label{tab:llm-sizes}
    \vspace*{0.2cm}
\end{table}

\myparagraph{Effect of LLM size}
We further investigate the effect of the LLM size, to verify that our improvements
still hold for smaller/larger LLMs, using the $7$B and $33$B versions of Vicuna.
Results are shown in Table~\ref{tab:llm-sizes}.
In general, the DR metric decreases (\silver: $33.3$; \dpr: $32.5$) using the $7$B version.
Further, we found that
the effect of scaling up the LLM to $33$B parameters
is negligible compared to enhancements on the retrieval side,
observing very similar results as for the $13$B version
(DR metric for \silver: $35.6$; \dpr: $34.3$).



\section{Related work}

\myparagraphnospace{Long-form question answering}
With the recent advances of LLMs~\cite{ouyang2022training, devlin2019bert, zheng2023judging} question answering has evolved beyond crisp and direct answers~\cite{yahya2013robust, bast2015more, sun2018open, saharoy2022question} toward supplying more in-depth and comprehensive passage-length responses.
There has been extensive research on LFQA
recently~\cite{nakano2021webgpt,
stelmakh2022asqa,
amplayo2022query, jiang2023active, fan2019eli5, su2022read, wang2022modeling, krishna2021hurdles, gao2023enabling, sun2023answering},
which has mostly built upon retrieval systems optimized for retrieving direct answers.
More details and discussion in the Appendix~\ref{app:related-work}.


\myparagraph{Weak supervision for training retrieval systems}
Obtaining training data for retrieval in QA has been a long-standing challenge~\cite{shen2023neural}.
The most common approach to obtain
training samples is to consider all
passages matching the direct answer as relevant~\cite{karpukhin2020dense, sachan2021end, joshi2017triviaqa}.
We extend this approach to the task of LFQA showing how to use both long-form and direct answers
for optimizing
retrieval toward contextual information.

\vspace*{0.2cm}
\section{Conclusion}
The retrieval of contextual information is often neglected in
LFQA,
while being an important ingredient for generating and grounding comprehensive long-form answers.
We investigate
techniques to obtain
training samples providing such contextual information
for
training re-ranking models.
We show that incorporating our re-rankers improves
retrieval and QA performance on a LFQA dataset,
yielding state-of-the-art performance on \asqa.
Notably, our method enhances groundedness of generated texts by 12.5\%,
which can reduce factual hallucinations in answers.
Experiments on \convmix show that our method, trained on \asqa, is able to generalize to an unseen ConvQA dataset.



\vspace*{0.7cm}
\myparagraph{Acknowledgements}
We thank Rexhina Blloshmi from Amazon AGI for useful inputs at various stages of this work.

Bill Byrne holds concurrent appointments as an Amazon Scholar and as Professor of Information Engineering at the University of Cambridge. This paper describes work performed at Amazon.

\clearpage

\section{Limitations}
\label{sec:limitations}
Our experimental setup with the \asqa dataset, which has reference long-form answers and short-form answers, allows us to investigate the duality of the LFQA and ConvQA tasks.
However, we only evaluated our approach on data that is publicly available.
We leave it for future work to run experiments with the approach in the wild.

In this work, we showed improvements on the retrieval side of a RAG pipeline based on Vicuna models of different sizes.
RAG pipelines based on other language model architectures might be affected differently by the enhanced retrieval recall provided by our approach, which is not investigated in this paper.

\section{Ethical considerations}
\label{sec:ethics}
We did not collect or release any private data or user data in this work.
All experiments are based on static datasets.

We make use of LLMs, which are known to generate factually incorrect texts or hallucinations.
In this work, we aim to enhance the grounding of long-form answers
by explicitly retrieving passages for contextual information.
Experiments indicate that with our approach the groundedness of answers can be improved,
which is a promising direction of reducing hallucinations in LLMs.

\bibliography{main}

\clearpage
\appendix
\begin{algorithm}[h]
    \caption{Pseudo-code for \silver approach}
    \label{alg:silver}
    \begin{algorithmic}[1]
        \newcommand{\LeftComment}[1]{\Statex \hspace*{-\algorithmicindent}#1}
        \newcommand{\comment}{\#\xspace}
        \LeftComment \textbf{Inputs:}
            \Statex $\mathcal{D}$: Dataset with questions, LFAs and DAs;
            \Statex $\mathcal{R}$: pre-trained BERT;
            \Statex $\mathcal{M}$: pre-trained causal LLM;
            \Statex \dpr: first-stage retriever;
            \Statex $k$: number of passages in context;
        \LeftComment \textbf{Outputs:}
            \Statex $\mathcal{R}_{SFT}$: fine-tuned re-ranker;
            \Statex $\mathcal{M}_{SFT}$: fine-tuned causal LLM;
        \LeftComment \rule{0.95\columnwidth}{0.4pt}
        \State {\color{blue}\comment Identify silver passages}
        \State $\mathcal{D}_P\gets \{\};$
        \ForAll{$(q, LFA, DA) \in \mathcal{D}$}
            \State {\color{blue}\comment First-stage retrieval (top-100)}
           \State $P_{DPR}\gets \dpr(q, 100);$
            \State {\color{blue}\comment Compute matching with LFA}
            \ForAll{$p \in P_{DPR}$}
                \State $score(p)\gets match(p, LFA);$ 
            \EndFor
            \State {\color{blue}\comment Compute silver passages $P^*$}
            \State $P^*_{DA}\gets \{p | DA \in p\};$ 
            \State $P^*_{LFA}\gets sort(\{p\}, score);$
            \State $P^*\gets top_k(P^*_{DA}, P^*_{LFA}, score);$
            \State $P^-\gets sample(P - P^*, 50);$
            \State {\color{blue}\comment Add to training data}
            \State $\mathcal{D}_P\gets \mathcal{D}_P \cup \{(q, P^*, P^-)\};$
        \EndFor

        \State {\color{blue}\comment Fine-tune re-ranking model}
            \State $\mathcal{R}_{SFT}\gets finetune(\mathcal{R}, \mathcal{D}_{P});$

        \State {\color{blue}\comment Fine-tune LLM}
        \State $\mathcal{D}_{LFA}\gets \{\}$;
        \ForAll{$(q, LFA, DA) \in \mathcal{D}$}
           \State $P_{DPR}\gets \dpr(q, 100)$;
           \State {\color{blue}\comment Apply re-ranker}
           \State $P_\mathcal{R}\gets \mathcal{R}_{SFT}(P_{DPR}, k)$;
            \State $\mathcal{D}_{LFA}\gets \mathcal{D}_{LFA} \cup \{(q, P_\mathcal{R}, LFA)\};$
        \EndFor
            \State $\mathcal{M}_{SFT}\gets finetune(\mathcal{M}, \mathcal{D}_{LFA});$
    \end{algorithmic}
\end{algorithm}





\section{Additional details on experiments}
\label{app:setup}

\myparagraph{Datasets}
{\asqa}
has $6{,}316$ ambiguous questions that originate from the Google search log~\cite{kwiatkowski2019natural, min2020ambigqa} paired with LFAs written by crowdworkers.
Every sample contains
a set of alternative question interpretations and a DA corresponding to each of them.
An example ambiguous question is \phrase{Who played bonnie in gone with the wind?},
with the two interpretations being
\phrase{Who played bonnie in the gone with the wind film?}
and
\phrase{Who played bonnie in the gone with the wind musical?}.
The corresponding DAs are \struct{Cammie King} and \struct{Leilah de Meza} in this case.
The dataset provides
one LFA for each question in the train set, and two LFAs for each question in the dev set, with an average of $64.8$ words per LFA (dev set).
Since the test set is hidden, we split the train set, using $95$\% for training and $5$\% for development, and the original dev set as our test set.
We used the official ASQA evaluation code\footnote{\url{https://github.com/google-research/language/tree/master/language/asqa}} to obtain the ROUGE-L, D-F1 and DR metrics. The dataset is licensed under an Apache License 2.0, thereby permitting its use for research purposes.

We use {\convmix} only for the evaluation of the models trained on the \asqa dataset,
since \convmix does not provide LFAs to train on.
We input only the first question from each conversation ($3{,}000$ questions, in total) and evaluate whether the generated LFA provides answers to the (first $4$) follow-up questions from the conversation.
The dataset is licensed under a CC BY 4.0, thereby permitting its use for research purposes.


\vspace*{0.2cm}
\myparagraph{Implementation details}
We implement \silver re-rankers as cross-encoders based on BERT models~\cite{devlin2019bert}\footnote{\url{https://huggingface.co/google-bert/bert-base-uncased}} with 110M parameters.
The input format is the following:
\struct{"[CLS] question [SEP] passage\_title [SEP] passage\_text"},
and the output is a scalar indicating the relevance of the respective passage.
We apply the re-ranker on top-$100$ \dpr results
(better performance than for top-$1{,}000$).
For training the re-ranker, for each question, we randomly sampled $50$ negatives (non-silver passages) along with $5$ positives (top-$5$ silver passages) from the top-$100$ \dpr passages.
We used AdamW as optimizer with a learning rate of $10^{-5}$, batch size of $16$, weight decay of $0.01$, and warm-up ratio of $0.04$.
Binary cross-entropy is used as loss function.

\vspace*{0.2cm}
\myparagraph{Comparison with state-of-the-art}
For this comparison, we used the full \asqa train set, same as in related work.
Note that the LLM here is trained for one epoch, without optimization on the dev set.
As \asqa is a rather small-scale dataset with only $4{,}353$ instances in the original train set,
the additional $5$\% (compared with our new split) make quite an impact on the LLM performance:
the DR metric improves from $35.5$ to $36.9$.

\vspace*{0.2cm}
\myparagraph{Computational costs}
Fine-tuning the BERT-based re-ranker took 50 minutes on an AWS EC2 P3 instance. The LLM fine-tuning (Vicuna 13B) took 110 minutes on an AWS EC2 P4 instance.

\section{Details on related work}
\label{app:related-work}
\myparagraph{Long-form question answering}
The ELI5 dataset~\cite{fan2019eli5} facilitated initial research on LFQA,
but due to evaluation problems~\cite{krishna2021hurdles}
recent work used the \asqa dataset with factoid long-form answers
for fairer comparison~\cite{stelmakh2022asqa, amplayo2022query, jiang2023active, sun2023answering}.
\citet{yehudai2024genie} propose Genie, an approach to create a synthetic LFQA dataset from Wikipedia, similar to ASQA.
The state-of-the-art methods for LFQA built upon retrieval systems
that are optimized for retrieving direct answers to questions~\cite{stelmakh2022asqa, krishna2021hurdles, sun2023answering, gao2023enabling}.

\citet{sun2023answering} proposed SIXPAQ, which constructs a database of potential questions
paired with their direct answers,
to augment the information obtained from the dense retriever~\cite{ni2022large}.
\citet{stelmakh2022asqa} used JPR~\cite{min2021joint}, an out-of-the-box re-ranker operating on top of \dpr results.
JPR is optimized for diversifying retrieval of direct answers, and thus for targeting ambiguous questions (the model is not publicly available).
This is different from our approach, which aims to retrieve both direct answers and contextual information for a question.
This allows our approach to produce more faithful long-form answers, enhancing the groundedness to the retrieval results.
Our experiments on a ConvQA dataset further show that our method works on non-ambiguous factoid questions without further training.

To the best of our knowledge, there is no existing work that optimizes the retrieval system towards contextual information, as required for generating comprehensive long-form answers.

\myparagraph{Iterative retrieval-augmentation}
A new line of work extends RAG pipelines~\cite{lewis2020retrieval} to multiple rounds of retrieval and generation~\cite{jiang2023active, shao2023enhancing, yao2022react}.

FLARE~\cite{jiang2023active} iteratively generates an upcoming next sentence, uses the generated sentence as query for retrieval,
and then generates the actual next sentence based on the retrieval results.
We compare against FLARE in Table~\ref{tab:asqa-sota}, and show that our approach can produce more suitable long-form answers.
IterRetGen~\cite{shao2023enhancing} first follows the standard RAG approach, but then iteratively
adds rounds of retrieval and generation to refine the initially generated text.
Their experiments are conducted on datasets with crisp short-form answers only.
ReAct~\cite{yao2022react} and Toolformer~\cite{schick2023toolformer} provide the LLM with specific actions that can trigger retrieval for a LLM-generated query.
The LLM itself can then generate relevant queries, and ground subsequent generations on the retrieval results.
ReAct is implemented based on in-context learning~\cite{brown2020language}, assuming strong instruction-following capabilities for the LLM.
Their main target are reasoning tasks, in which the LLM interacts with an environment to predict a sequence of actions.
The Toolformer makes use of LLM-perplexity for identifying relevant calls of tools (such as retrieval with a specific query).
We investigate the underlying idea in this work (Table~\ref{tab:main-res}) for identifying relevant silver passages.

Note that these approaches, by design, employ multiple rounds of generation and retrieval, making them
intractable in many real-world scenarios in which users expect an answer within a few seconds (at most).
Further, iterative retrieval and generation can lead to extremely long prompts, as previous retrieval and generation results are often retained as context for the LLM.

Our approach aims to retrieve the most relevant information 
in \textit{one shot}, and then ground the answer on these one-time retrieval results.

\end{document}